\documentclass{article}
\usepackage{color}
\usepackage{spconf,amsmath,graphicx}
\usepackage[hidelinks]{hyperref}
\usepackage{float}
\usepackage{booktabs}
\usepackage{tabularx}
\usepackage{array}
\usepackage[normalem]{ulem}
\newcolumntype{C}{>{\centering\arraybackslash}X}

\title{Progressively Refining SAM for Accurate Phytoplankton Segmentation}
\name{Jialin Li \quad Jinkun Jiang \quad Yuezun Li$^{*}$\thanks{* Corresponding author} \quad Junyu Dong}
\address{School of Computer Science and Technology,  Ocean University of China, Qingdao, China}
\begin{document}
\maketitle

\begin{abstract}
Accurate phytoplankton segmentation is essential for ecological monitoring, yet it remains largely underexplored due to the subtle visual characteristics of phytoplankton and the scarcity of pixel-level annotations. To address these challenges, we propose \textit{PhySeg}, a SAM-based segmentation framework equipped with a lightweight refinement module that progressively restores fine-grained details through multi-scale feature fusion. In addition, we construct a phytoplankton segmentation dataset by manually annotating instance-level masks for an existing dataset, providing a valuable resource for future research. Experimental results demonstrate that PhySeg improves segmentation accuracy, validating the effectiveness of the proposed module. Code is available at \url{https://github.com/lijialin404/PhySeg}.

\vspace{-0.5em}
\end{abstract}
\vspace{-0.5em}

\section{Introduction}
\label{sec:introduction}
Phytoplankton underpin marine primary production and food webs and play an important role in the global carbon cycle~\cite{Falkowski1998}. Characterizing their morphology and behavior is therefore essential for understanding marine ecosystems. Several studies have focused on phytoplankton classification~\cite{eerola2024survey,taheriashtiani2026review}, supported by public benchmarks such as WHOI-Plankton~\cite{orenstein2015whoi}. Beyond classification, detection and tracking methods~\cite{huo2025enhancing,zhang2020lightweight,yu2025phytracker} enable the spatial localization and temporal association of individual organisms, supported by datasets such as PMID2019~\cite{li2020pmid} and PMOT2023~\cite{PMOT2023}.

\begin{figure*}[!t]
    \centering
    \includegraphics[width=\textwidth]{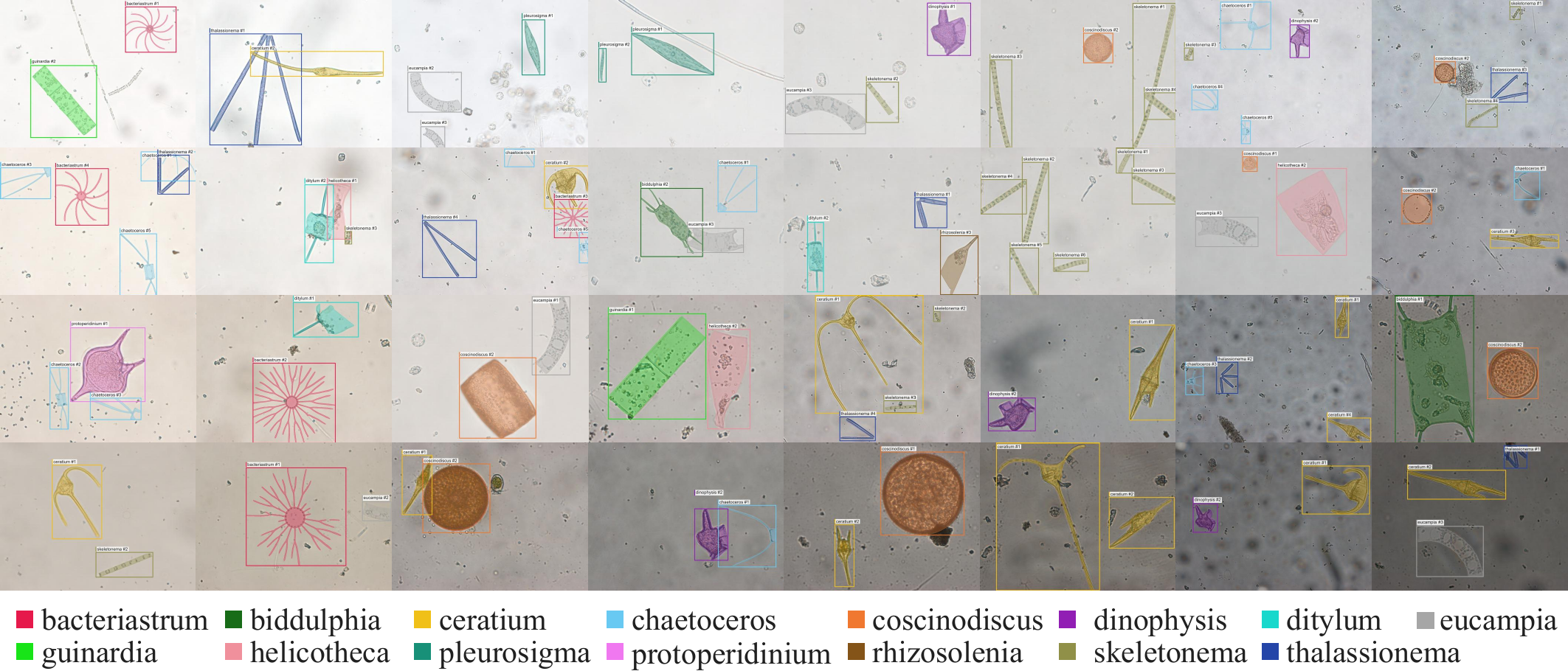}
    \vspace{-16pt}
    \caption{Visual examples of the phytoplankton segmentation dataset.}
    \label{fig:figure_sample}
    \vspace{-0.4cm}
\end{figure*}

Despite this progress, category labels, bounding boxes, and trajectories provide limited information about the precise morphology of individual organisms. Therefore, pixel-level segmentation for phytoplankton is required to support more detailed ecological analysis~\cite{sonnet2022morphology}. However, phytoplankton segmentation remains comparatively underexplored, with two major challenges. 
First, general-purpose segmentation models, including the Segment Anything Model (SAM)~\cite{kirillov2023sam}, are not specifically designed for phytoplankton. Substantial variations in morphology and scale, together with slender structures, low contrast, and ambiguous boundaries (see examples in Fig.~\ref{fig:figure_sample}), make it difficult to recover fine details reliably~\cite{eerola2024survey,li2020pmid,sonnet2022morphology}. Second, adapting these models requires high-quality segmentation annotations, which remain scarce in existing phytoplankton datasets. Recent surveys likewise identify limited annotated data and the need for segmentation-oriented methods as persistent challenges in phytoplankton image analysis~\cite{eerola2024survey,taheriashtiani2026review}.

In this paper, we address these challenges through two contributions. {1) Methodology.} We introduce a dedicated phytoplankton segmentation framework built on SAM2~\cite{ravi2025sam2}, named \textit{PhySeg}. Drawing on its strong general-purpose segmentation capabilities, we adapt SAM2 to phytoplankton microscopy and develop an add-on refiner that integrates multi-scale features to progressively recover fine structural details. {2) Dataset.} We extend PMID2019 with 11990 manually annotated instance masks across 9824 images. Unlike general objects, annotating phytoplankton requires substantial manual effort due to their slender structures and intricate boundaries. Despite its modest scale, this dataset provides a valuable resource for developing and evaluating phytoplankton segmentation methods.
Experiments on this dataset demonstrate the effectiveness of the proposed framework, establishing a baseline for phytoplankton segmentation.

\begin{figure*}[!t]
    \centering
    \includegraphics[width=\textwidth]{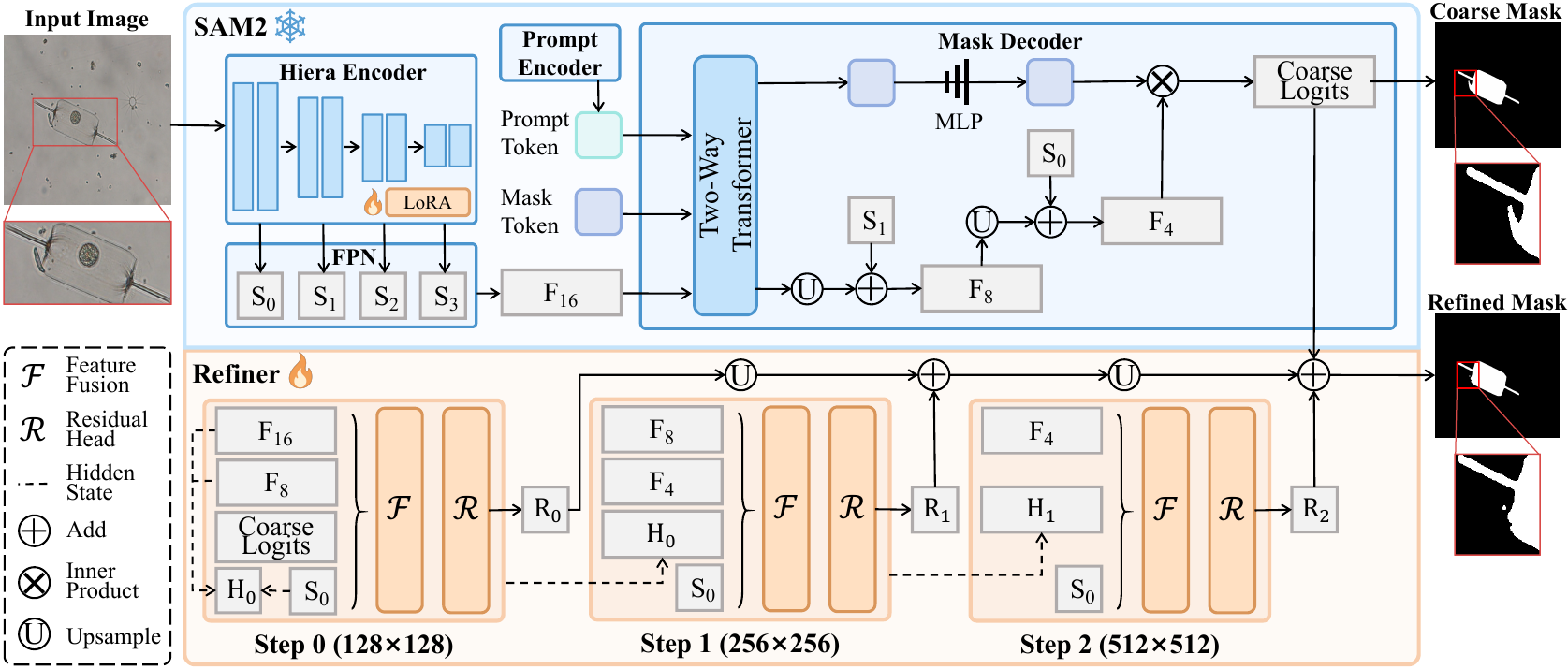}
    \vspace{-16pt}
    \caption{Detailed structure of the proposed PhySeg.}
    \label{fig:figure_ours}
    \vspace{-0.4cm}
\end{figure*}
\begin{figure}[!t]
    \includegraphics[width=\columnwidth]{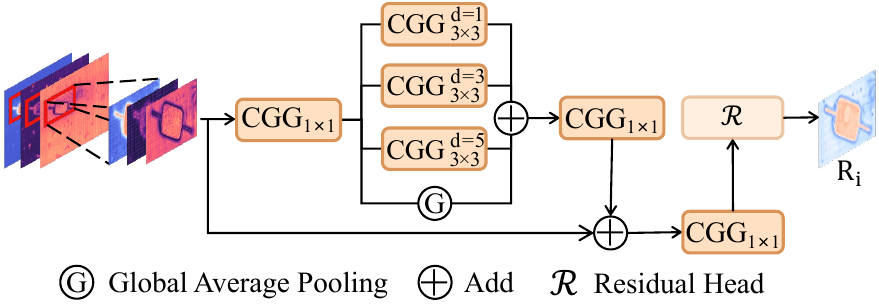}
    \vspace{-16pt}
    \caption{Overview of the Feature Fusion module. $\mathrm{CGG}_{k\times k}^{d}$ denotes a $k\times k$ convolution with dilation rate $d$, followed by GroupNorm and GELU. The dilation rate is 1 when omitted.}
    \label{fig:figure_detailed}
    \vspace{-0.4cm}
\end{figure}
\section{Method}
\label{sec:method}
PhySeg is built upon SAM2 and introduces a lightweight refiner to progressively recover fine-grained structures. Specifically, given an input image and a box prompt, SAM2 first produces coarse logits, which are then progressively refined by fusing them with multi-scale features. To adapt SAM2 to phytoplankton segmentation, we apply LoRA~\cite{hu2022lora} to the image encoder. The overall architecture is illustrated in Fig.~\ref{fig:figure_ours}.

\smallskip
\noindent\textbf{Progressive Refinement.} 
In SAM2, coarse logits are directly upsampled and thresholded to obtain the final mask. However, upsampling alone does not explicitly correct errors in the coarse predictions.
This naturally raises a question: can we refine the coarse predictions during upsampling?

Motivated by this idea, a Progressive Refinement strategy is introduced, which refines coarse logits through multiple steps, rather than relying on a single step to address various errors. Each step consists of two small convolutional modules, named Feature Fusion and Residual Head. The Feature Fusion module aggregates the input features at the current step, while the Residual Head module takes the fused features as input and predicts the residual logits. The residual logits are added to the coarse logits step by step to progressively refine the results.

\smallskip
\noindent\textbf{Structure Design.}
SAM2 inherently encodes semantic and spatial information at multiple scales. We therefore reorganize these intermediate features as complementary cues for mask refinement.
Specifically, we use three feature maps from the decoding process, denoted as $F_{16}$, $F_{8}$, and $F_{4}$, with spatial resolutions of $1/16$, $1/8$, and $1/4$ of the input image, respectively. 

While $F_{16}$ provides rich semantic information, its limited spatial detail compromises the initial representation on which subsequent corrections rely. We therefore introduce $F_8$ at Step 0 to incorporate finer spatial cues from the outset, providing a stronger foundation for subsequent refinement.
At Step 1, we introduce $F_{4}$ for finer spatial details while retaining $F_8$ to provide complementary structural context.
After this feature transition in the first two steps, Step 2 directly uses $F_{4}$ to further handle finer local errors. 
In this way, the refinement process gradually shifts from correcting the overall structure to improving local details.

In addition, the updated logits at each step, together with the fused features, serve as the hidden state $H_i$, carrying forward the refined features and current mask prediction and guiding the next step to achieve a balance between area and boundaries. 
At Step 0, an additional Feature Fusion module aggregates context from the input features before they are combined with the coarse logits to construct the initial hidden state $H_0$.
Besides, the shallow encoder feature $S_0$ is introduced at every step to provide local spatial details.

As shown in Fig.~\ref{fig:figure_detailed}, inspired by the ASPP module~\cite{chen2018deeplab}, we design the Feature Fusion module to address the need for both local and global information during refinement. 
Input features are first aligned via Region of Interest (ROI) and then processed by three parallel convolutional branches with different dilation rates to capture local information, while Global Average Pooling is added to provide a shared global reference for refinement at different spatial locations.
The fused features are further fed into the Residual Head $\mathcal{R}$, which consists of a $3\times3$ convolution followed by a $1\times1$ convolution, to predict the residual logits for correction at each step. 

\smallskip
\noindent\textbf{Training Objective.}
The refined prediction is supervised using a combination of focal loss~\cite{lin2017focal} and Dice loss~\cite{milletari2016vnet}: $\mathcal{L}=\mathcal{L}_{\mathrm{Focal}}+\mathcal{L}_{\mathrm{Dice}}$.

\begin{table}[!t]
    \centering
    \caption{Category statistics of the dataset.}
    \label{tab:dataset_stats}
    \begin{tabular*}{\columnwidth}
        {@{\extracolsep{\fill}}c c c c@{}}
        \toprule
        \textbf{ID} & \textbf{Genus} &
        \textbf{Images} & \textbf{Instances} \\
        \midrule
         1 & bacteriastrum   &   677 &   703 \\
         2 & biddulphia      &   356 &   360 \\
         3 & ceratium        &  2267 &  2332 \\
         4 & chaetoceros     &  1095 &  1184 \\
         5 & coscinodiscus   &  1123 &  1140 \\
         6 & dinophysis      &   413 &   417 \\
         7 & ditylum         &    99 &   100 \\
         8 & eucampia        &  1446 &  1514 \\
         9 & guinardia       &   857 &   869 \\
        10 & helicotheca     &   369 &   374 \\
        11 & pleurosigma     &   228 &   237 \\
        12 & protoperidinium &   105 &   105 \\
        13 & rhizosolenia    &    77 &    77 \\
        14 & skeletonema     &  1053 &  1141 \\
        15 & thalassionema   &  1389 &  1437 \\
        \bottomrule
    \end{tabular*}
    \vspace{-0.4cm}
\end{table}

\begin{table*}[!t]
    \caption{
        Quantitative performance comparison of different methods on the test set. ``Box'' indicates that a bounding box is used as the model prompt, whereas ``Mask'' indicates that the SAM2+LoRA prediction is provided as the initial mask.
    }
    \label{tab:comparison}
    \begin{tabularx}{\textwidth}{
        @{}
        c|C|C|CCCCC
        @{}
    }
        \toprule
        \textbf{Method}
        & \textbf{Venue}
        & \textbf{Prompt}
        & \textbf{mIoU$\uparrow$}
        & \textbf{Dice$\uparrow$}
        & \textbf{BF1$\uparrow$}
        & \textbf{BF2$\uparrow$}
        & \textbf{BF4$\uparrow$} \\
        \midrule
        CascadePSP$^{*}$~\cite{cheng2020cascadepsp}
        & CVPR'20
        & Mask
        & 87.08
        & 92.77
        & 53.45
        & 72.41
        & 90.57 \\
        BPR$^{*}$~\cite{tang2021bpr}
        & CVPR'21
        & Mask
        & 88.51
        & 93.53
        & 59.61
        & 77.99
        & 92.88 \\
        CRM$^{*}$~\cite{shen2022crm}
        & CVPR'22
        & Mask
        & 68.20
        & 76.08
        & 34.94
        & 49.04
        & 65.39 \\
        SAM2$^{*}$~\cite{ravi2025sam2}
        & ICLR'25
        & Box
        & 81.03
        & 87.82
        & 42.86
        & 60.29
        & 80.46 \\
        HQ-SAM2$^{*}$~\cite{ke2023hqsam}
        & NeurIPS'23
        & Box
        & 82.99
        & 89.35
        & 47.57
        & 65.83
        & 84.64 \\
        \midrule
        CascadePSP~\cite{cheng2020cascadepsp}
        & CVPR'20
        & Mask
        & 89.72
        & 94.37
        & 62.69
        & 79.85
        & 93.36 \\
        BPR~\cite{tang2021bpr}
        & CVPR'21
        & Mask
        & \underline{91.73}
        & \underline{95.60}
        & \underline{67.79}
        & 83.51
        & 94.93 \\
        CRM~\cite{shen2022crm}
        & CVPR'22
        & Mask
        & 67.19
        & 75.03
        & 36.00
        & 51.55
        & 68.36 \\
        SAM2+LoRA~\cite{hu2022lora}
        & ICLR'22
        & Box
        & 91.20
        & 95.30
        & 65.50
        & 82.52
        & 94.86 \\
        HQ-SAM2~\cite{ke2023hqsam}
        & NeurIPS'23
        & Box
        & \underline{91.73}
        & \underline{95.60}
        & 67.52
        & \underline{83.71}
        & \textbf{95.31} \\
        \textbf{PhySeg (Ours)}
        & --
        & Box
        & \textbf{91.87}
        & \textbf{95.68}
        & \textbf{68.32}
        & \textbf{84.02}
        & \underline{95.19} \\
        \bottomrule
    \end{tabularx}
    \vspace{-0.4cm}
\end{table*}

\section{Phytoplankton Segmentation Dataset}
\label{sec:Dataset}
Few public datasets provide pixel-level annotations for phytoplankton segmentation. To address this gap, we construct the Phytoplankton Segmentation Dataset (PhySegD) by manually annotating ground-truth masks for the PMID2019 dataset.

\textit{Creating PhySegD is particularly labor-intensive} despite its moderate scale. In bright-field microscopy images, halos surrounding phytoplankton are often difficult to distinguish from the boundaries of semi-transparent cells, necessitating multiple rounds of expert review. 
Moreover, impurities with textures similar to those of phytoplankton may interfere with or obscure cell boundaries, further complicating accurate mask annotation. Representative samples from different genus-level categories and their annotations are shown in Fig. \ref{fig:figure_sample}, while Table \ref{tab:dataset_stats} reports the numbers of images and instances across the 15 categories. 
Overall, PhySegD contains 9824 images and 11990 instances, divided into 6880 training images with 8378 instances, 1968 validation images with 2395 instances, and 976 test images with 1217 instances.

\section{Experiments}
\label{sec:Exp}
\noindent\textbf{Experimental Setup.}
We evaluate PhySeg on the proposed PhySegD dataset and compare it with representative mask-refinement methods, including CascadePSP~\cite{cheng2020cascadepsp}, BPR~\cite{tang2021bpr}, CRM~\cite{shen2022crm}, and HQ-SAM2~\cite{ke2023hqsam}. We report mean Intersection over Union (mIoU) and Dice scores to evaluate region accuracy, together with boundary F-scores at thresholds of 1, 2, and 4 pixels (BF1, BF2, and BF4) to assess boundary quality~\cite{perazzi2016davis}. Higher values indicate better performance for all metrics. \textit{Among these metrics, mIoU serves as the primary measure of overall segmentation accuracy.}

\smallskip
\noindent\textbf{Implementation Details.}
During training, we apply random horizontal and vertical flips, rotations by multiples of \(90^{\circ}\), and color jittering. The same geometric transformations are applied consistently to the images, masks, and bounding boxes.
Training is conducted in two stages. First, we adapt SAM2 to phytoplankton images using LoRA. The model is optimized for 15 epochs using AdamW with a learning rate of \(2\times10^{-5}\), a weight decay of 0.01, and a batch size of 1. Second, we freeze SAM2 and its LoRA parameters and train only the refiner for 15 epochs. We use AdamW with an initial learning rate of \(5\times10^{-5}\), a weight decay of 0.01, and a batch size of 1. The learning rate follows a cosine annealing schedule with a minimum value of \(10^{-6}\). 

\begin{figure}[!t]
    \includegraphics[width=\columnwidth]{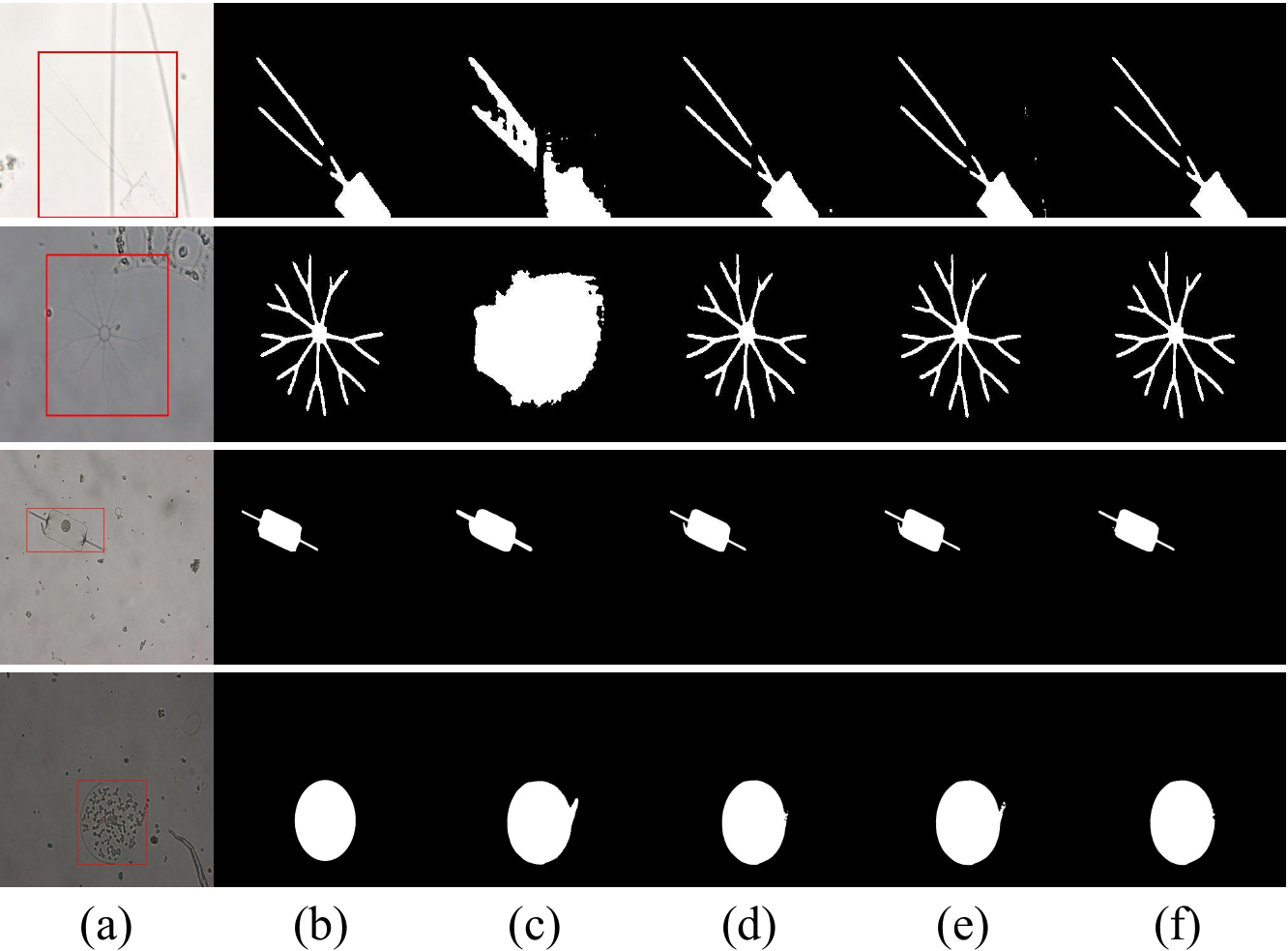}
    \vspace{-16pt}
    \caption{Visual examples of our method and others:  (a) original image, (b) ground truth, (c) SAM2, (d) SAM2+LoRA, (e) HQ-SAM2, and (f) PhySeg (Ours).}
    \vspace{-0.4cm}
    \label{fig:figure_comp}
\end{figure}

\smallskip
\noindent\textbf{Comparison Results.}
Table~\ref{tab:comparison} compares PhySeg with existing segmentation and refinement methods. 
The upper part with the superscript \(^{*}\) denotes models evaluated using their official pretrained weights without fine-tuning. 
The lower part reports the results of these models after fine-tuning on our dataset. For CRM, BPR, and CascadePSP, we fine-tune the official pretrained models using their recommended training configurations. For HQ-SAM2, we reimplement the HQ module ourselves, strictly following its instructions. PhySeg achieves the best mIoU, BF1, BF2, and Dice scores, reaching 91.87\%, 68.32\%, 84.02\%, and 95.68\%, respectively. It also outperforms the mask-based refinement methods despite requiring only a box prompt. Fig.~\ref{fig:figure_comp} further shows that PhySeg better preserves slender structures and corrects inaccurate regions in the initial masks.

BF1 and BF2 require the predicted boundary to be closer to the real boundary, while BF4 uses a wider tolerance. Compared with HQ-SAM2, our method achieves improvements in mIoU, Dice, BF1, and BF2, with a slight decrease of 0.12 percentage points in BF4. 
Overall, these results show that PhySeg achieves a better balance between boundary refinement and region correction, which does not contradict the fact that there is still room for improvement in BF4.

\smallskip
\noindent\textbf{Ablation Study.}
\begin{table}[!t]
\caption{Ablation study on the val-set.}
\label{tab:ablation}
\setlength{\tabcolsep}{3.2pt}
\renewcommand{\arraystretch}{1.10}
    \begin{tabular*}{\columnwidth}{
        @{\extracolsep{\fill}} c|ccccc @{}
    }
        \toprule
        \textbf{Method}
        & \textbf{mIoU$\uparrow$}
        & \textbf{Dice$\uparrow$}
        & \textbf{BF1$\uparrow$}
        & \textbf{BF2$\uparrow$}
        & \textbf{BF4$\uparrow$} \\
        \midrule
        PhySeg
        & \textbf{92.07} & \textbf{95.80}
        & \underline{67.75} & \underline{84.11} & 95.47 \\
        w/o $H$
        & 91.97 & 95.74
        & \textbf{67.92} & \textbf{84.35} & \textbf{95.61} \\
        w/o $S_0$
        & \underline{92.03} & \underline{95.77}
        & 67.65 & 83.95 & 95.49 \\
        w/o Step 0
        & 92.00 & 95.76
        & 67.45 & 84.04 & \underline{95.55} \\
        w/o Step 0 \& Step 1
        & 91.96 & 95.74
        & 67.36 & 83.88 & 95.45 \\
        \bottomrule
    \end{tabular*}
    \vspace{-0.4cm}
\end{table}
Table~\ref{tab:ablation} evaluates the contributions of the feature inputs and progressive refinement steps. $H$ denotes the hidden state transferred across steps; $S_0$ denotes the shallow feature extracted from Hiera Stage 0. As shown in Fig.~\ref{fig:figure_ours}, Steps 0, 1, and 2 denote the progressive procedure of refinement. We transfer $H$ across steps to find a balance between area refinement and boundary refinement. As shown by the ablation study, adding $H$ leads to an increase in mIoU and Dice while causing a slight decrease in BF, which is consistent with the design of $H$. 

In addition, removing $S_0$, Step 0, or Step 0 and Step 1 consistently leads to performance degradation across multiple metrics. Since the original SAM2 already provides a relatively strong segmentation baseline, the resulting changes are not necessarily large in magnitude. Moreover, these components contain partially overlapping feature information, which may further limit the apparent gain contributed by each individual component. However, the overall trend observed in the ablation results is consistent with our design motivation: these features are introduced as refinement prompts to progressively guide the segmentation results.

\section{Conclusion}
\label{sec:Conclusion}
In this work, we propose PhySeg, a phytoplankton segmentation method built upon SAM2 and equipped with a newly designed refiner module. By progressively fusing multi-scale features, the refiner recovers missing details and corrects inaccurate regions in the initial segmentation, thereby improving the delineation of fine phytoplankton structures. In addition, we construct a phytoplankton segmentation dataset by manually annotating ground-truth masks for the existing PMID2019 dataset. Experimental results demonstrate the effectiveness of the proposed method.
\newpage
\bibliographystyle{IEEEbib}
\bibliography{refs}
\end{document}